\documentclass[10pt,twocolumn,letterpaper]{article}

\usepackage{cvpr}              

\usepackage[accsupp]{axessibility}  
\usepackage{booktabs} %
\usepackage[dvipsnames]{xcolor}
\definecolor{cvprblue}{rgb}{0.21,0.49,0.74}
\usepackage[pagebackref,breaklinks,colorlinks,citecolor=cvprblue]{hyperref}

\def\paperID{11} 
\def\confName{CVPRW PhysHuman\xspace}
\def\confYear{2026\xspace}

\usepackage[ruled]{algorithm2e} %

\SetAlFnt{\small}
\SetAlCapFnt{\small}
\SetAlCapNameFnt{\small}
\SetAlCapHSkip{0pt}

\title{GeomHair: Reconstruction of Hair Strands from Colorless 3D Scans}

\author{
  Rachmadio Noval Lazuardi$^{*1}$ \quad Artem Sevastopolsky$^{*1}$ \quad Egor Zakharov$^2$ \\
  Matthias Nie{\ss}ner$^1$ \quad Vanessa Sklyarova$^{3,2}$ \\[12pt]
  $^1$ Technical University of Munich \quad $^2$ ETH Z\"urich \quad $^3$ Max Planck Institute for Intelligent Systems
}

\def\S{\mathbf{S}} %
\def\p{\mathbf{p}} %

\def\Zgeom{\mathbf{T}} %
\def\zgeom{\mathbf{z}} %
\def\x{\mathbf{x}} %
\begin{document}

\twocolumn[{%
    \renewcommand\twocolumn[1][]{#1}%
    \vspace{-0.75cm}
    \maketitle
    \vspace{-0.5cm}
    \includegraphics[width=1\textwidth,trim={0cm 0cm 0cm 0cm},clip]{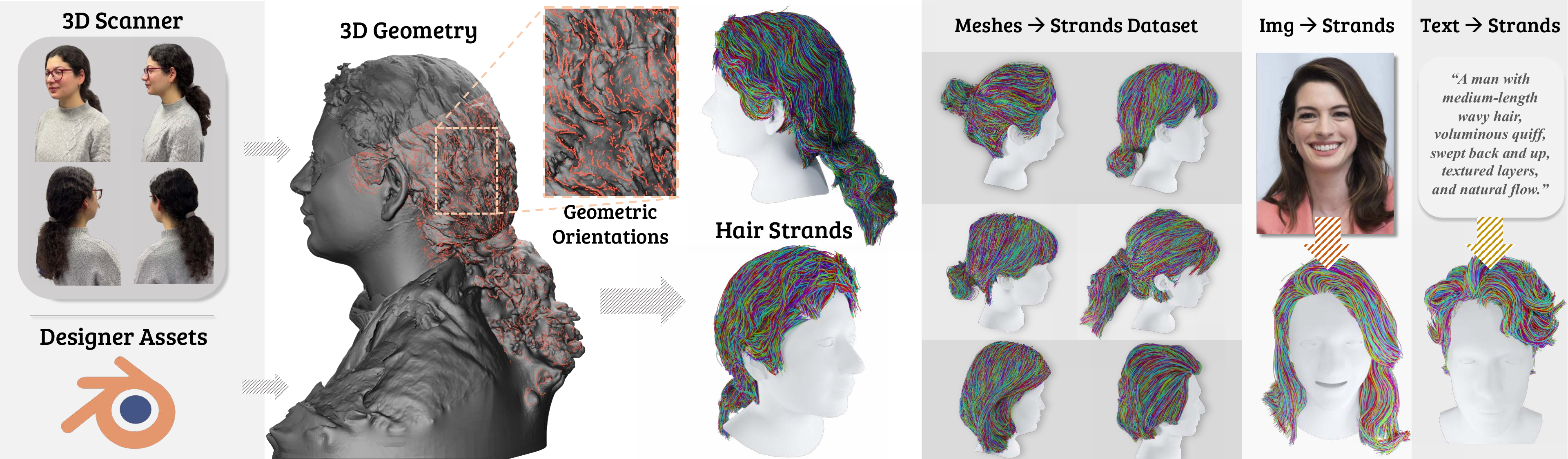}
    \vspace{-0.6cm}
    \captionof{figure}{
    We present \emph{GeomHair} -- a method for reconstructing complete hair strands representations from 3D scans that can be obtained from various sources, such as handheld 3D scanners, designer assets, and others. 
    Our method extracts information about guiding sharp features directly from the scan geometry by employing a combination of 3D and 2D orientation detectors and fits strands with a diffusion prior conditioned on the scan-specific prompt embedding.
    We also provide a dataset of reconstructed 3D assets obtained using our method from the meshes produced by a structured light scanner.
    }
    \vspace{0.4cm}
    \label{fig:overview}
}]

\begin{abstract}
    \vspace{-0.2cm}
We propose a novel method that reconstructs hair strands directly from colorless 3D scans by leveraging multi-modal hair orientation extraction.
Hair strand reconstruction is a fundamental problem in computer vision and graphics, essential for high-fidelity digital avatar synthesis, animation, and AR/VR applications.
However, accurately recovering hair strands from raw scan data remains challenging due to the complex and fine-grained structure of human hair, and none of the existing methods operate on colorless 3D geometry alone.
To address this gap, our method directly identifies sharp surface features on the scan and estimates strand orientation using a neural 2D line detector applied to the renderings of scan shading.
Additionally, we incorporate a diffusion prior trained on a diverse set of synthetic hair scans, refined with a noise schedule, and adapted to the reconstructed contents via a scan-specific text prompt.
We demonstrate that this combination of supervision signals enables accurate reconstruction of both simple and intricate hairstyles from geometry alone.
By enabling strand extraction from 3D scans, we compile Strands400, the largest publicly available dataset of hair strands with detailed surface geometry extracted from real-world data, comprising reconstructions from 400 subjects' scans. Strands400 enables training data-driven generative models for downstream tasks such as image-to-strands and text-to-strands.
Moreover, our method applies to designer mesh assets, supporting a practical CG workflow where artists model hair as meshes and need strand-level representations for simulation and rendering. All code and data will be released for research purposes on \url{https://seva100.github.io/GeomHair/}.

\end{abstract}



\section{Introduction}
\label{sec:intro}



Reconstructing hair as 3D curves is a highly challenging problem at the intersection of computer vision and computer graphics, yet it is essential for creating high-fidelity digital avatars.
Additionally, hair strands remain a critical component in applications such as telepresence, motion capture, gaming, and virtual character design.
Unlike surface-only reconstructions, modeling individual hair strands enables physically accurate motion under head movement and external forces.
%
Furthermore, explicitly representing the internal hair volume through strands is critical for achieving realistic, physically-based hair rendering.
While modern video generation models~\cite{wan2025, yang2024cogvideox, kong2024hunyuanvideo, wiedemer2025video} can animate hair with unprecedented realism, creating and reconstructing complete head models, characters, or avatars still requires explicit strand-based representations.

A key bottleneck for advancing strand-based hair modeling is the scarcity of large-scale, diverse datasets of real hair strands.
Training data-driven generative models demands a huge amount of strand-level hairstyles, yet existing datasets rely on extensive augmentation~\cite{groomgen, haar} or costly, handcrafted parametric hair models~\cite{perm}, limiting both scale and diversity.
Besides, benchmarking both strands reconstruction models and generative models for strands requires a significant number of real strands collections to be publicly available.
As a result, this data gap directly constrains the quality and generalization of downstream models.

Creating large amounts of real strands collections not only would require a extremely tedious designing work, but could also be suboptimal when representing imperfect hairstyles plausibly.
Besides, designing such a hairstyle would require not only modeling its geometry but also its material properties to correctly model resting hair under gravity force. 
An alternative way of collecting such data would be through auto-annotation.
For this, existing prior art on reconstruction could be leveraged, in which a variety of methods to infer the location and orientation of strands from various input modalities have been proposed, ranging from controlled multi-view captures under light stages~\cite{Nam2019StrandAccurateMH, McGuire2021HumanHI, neuralstrands, Wang2022NeuWigsAN, Wang2021HVHLA} to more accessible, hand-held RGB videos acquired by a smartphone~\cite{neuralhaircut, luo2024gaussianhair, GH, MH, GroomCap}.
However, all of these methods fundamentally depend on high-quality multi-view color information, which is rarely publicly available at scale.

This reliance prevents the existing methods from exploiting a vast and growing source of 3D data: colorless surface scans.
Advances in 3D scanning have produced large, publicly available collections of highly detailed head meshes~\cite{giebenhain2023learning}, yet no current method can extract strands from such data.
Similarly, 3D hair asset designers craft hairstyles starting from a base mesh and guide strands (see grooming guides for Houdini~\cite{sidefx_houdini_intro_grooming,sidefx_houdini_realistic_hair} and 3ds Max~\cite{youtube_3dsmax_hair_tutorial}) and need to convert these meshes into dense strand-level representations for simulation and rendering.
Furthermore, the rapid progress of 3D mesh generation from text and images~\cite{meshy2025} opens the door to creating hair strands from these simple input modalities, enabling text-to-strands and image-to-strands pipelines.
Together, these trends call for a method that can reconstruct hair strands directly from 3D geometry -- without color -- thereby unlocking scalable dataset creation from existing scan collections, designer assets, and generated meshes alike.

To this end, we introduce GeomHair, the first method to reconstruct hair strands directly from colorless scan geometry. Our approach identifies prominent guiding features directly from the 3D surface by combining two complementary orientation sources -- geometric cues from mesh ridges and ravines via Crest Lines~\cite{yoshizawa2005fast}, and 2D cues from multi-viewpoint shading with neural orientation detection~\cite{soria2023tiny} -- with a text-conditioned diffusion prior trained on HAAR synthetic hair data~\cite{haar}, refined via an optimized noise schedule. We further condition the prior on text embeddings generated by a vision-language model~\cite{liu2023llava} from scan renderings, enabling flexible handling of diverse hairstyles. By operating on geometry alone, GeomHair enables scalable strand extraction: we apply it to an extended collection of 3D head scans to compile Strands400, the largest publicly available real-world strand dataset with 400 subjects, directly addressing the data bottleneck that limits current generative hair models. The result is high-quality strand reconstruction from colorless geometry, as illustrated in Fig.~\ref{fig:overview}.

To summarize, our contributions are as follows:
\begin{itemize}
    \itemsep0pt
    \item We introduce \textit{GeomHair}, the first method for reconstructing hair strands directly from colorless 3D scans.
    \item Our method enables the creation of \textit{Strands400} dataset, the largest publicly available real-world strand dataset of 400 subjects curated from high-quality scans.
    \item We show the practicality of our method by creating strands from off-the-shelf text-to-mesh and image-to-mesh pipelines~\cite{meshy2025}, enabling strand-level workflows on top of mesh generators.
    \item Our method also supports a practical CG workflow in which artists craft hairstyles from a base mesh and obtain dense strand-level output for simulation and physically based rendering.
\end{itemize}

\section{Related Work}
\label{sec:related}

\begin{figure}[t!]
    \vspace{-0.1cm}
    \centering
    \includegraphics[clip,trim={0cm 6cm 0cm 0cm},width=\linewidth]{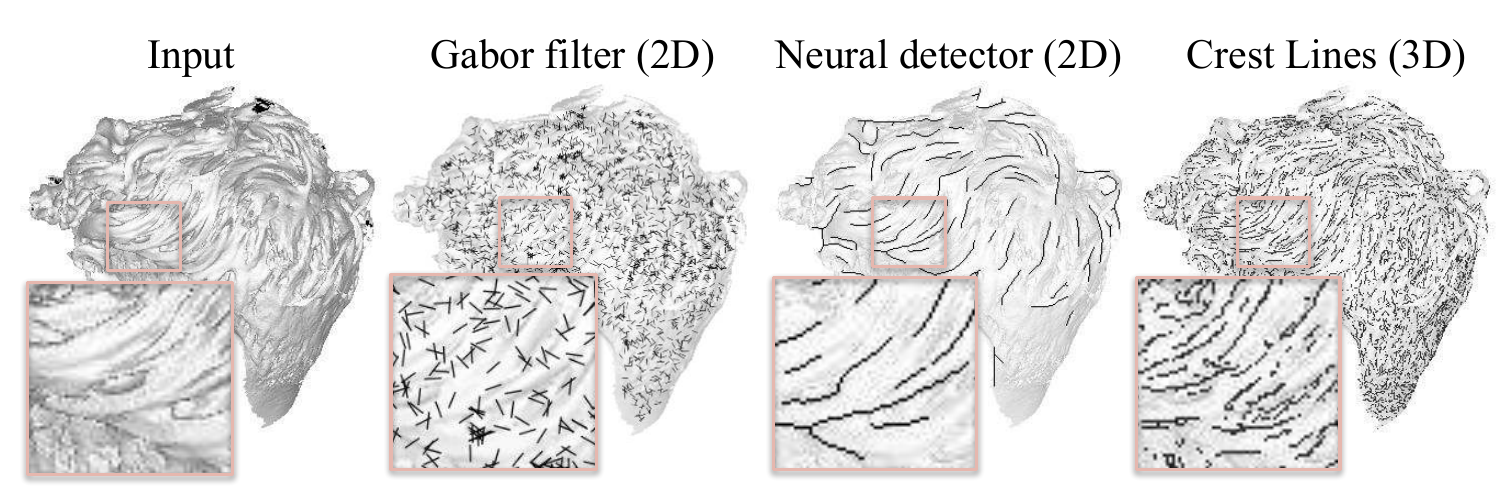}
    \vspace{-0.65cm}
    \caption{
    Commonly used orientations detectors, such as Gabor filter~\cite{Paris2004CaptureOH}, yield noisy results, especially when applied to grayscale shading.
    In our work, we employ a 2D neural orientation detector TEED~\cite{soria2023tiny} and 3D Crest Lines detector~\cite{yoshizawa2005fast} that produce less noisy and more accurate directions.
    }
    \label{fig:ablation_filters}
    \vspace{-0.4cm}
\end{figure}

\begin{figure*}
    \includegraphics[width=1\textwidth,trim={0.1cm 7.2cm 0.5cm 0cm},clip]{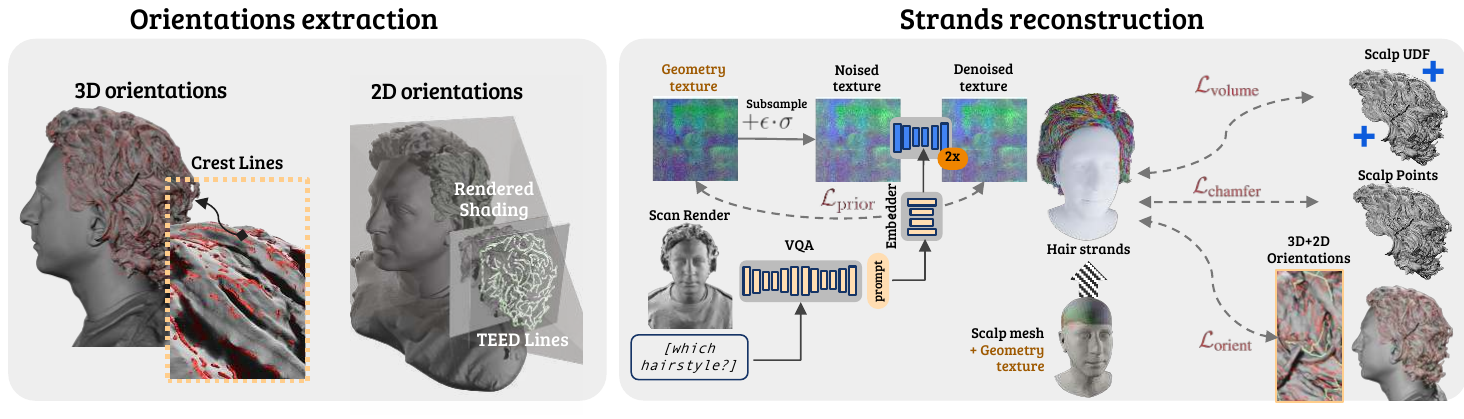}
    \vspace{-0.8cm}
    \captionof{figure}{
    \textbf{Overview of our GeomHair framework.} Our method consists of two main stages: orientations extraction (left) and strands reconstruction (right). In the orientation extraction stage, we extract complementary orientation signals by combining 3D orientations from crest lines with 2D orientations obtained from TEED features applied to rendered shading of the scans. During reconstruction, we optimize a geometry texture to generate hair strands while enforcing multiple constraints: the orientation loss ($\mathcal{L}_\text{orient}$) ensures strand growth aligns with our extracted 3D+2D orientation field, volume loss ($\mathcal{L}_\text{volume}$) keeps strands near the surface, and Chamfer distance ($\mathcal{L}_\text{chamfer}$) promotes uniform coverage of the hair volume. To enhance realism, we additionally incorporate a diffusion prior ($\mathcal{L}_\text{prior}$) conditioned on hairstyle descriptions generated by a VQA model analyzing the input scan.
    }
    \vspace{-0.2cm}
    \label{fig:method}
\end{figure*}

\noindent\textbf{2D and 3D Line Detection.}
3D hair reconstruction heavily relies on line detection, as hair strands can be effectively approximated by 3D polylines.
Various 2D line detectors have been explored for hair reconstruction, including edge detection filters such as Canny~\cite{Canny1986ACA} and Gabor~\cite{Paris2004CaptureOH}, with the latter widely adopted in recent methods~\cite{hu2015single, hairnet, neuralhdhair, neuralstrands, GH, MH, luo2024gaussianhair, GroomCap, deepmvshair}.
However, despite yielding improved results on RGB images, Gabor-based line orientation maps often fail under low-light or strongly directional lighting conditions.
Moreover, Gabor filtering heavily depends on the reflectance properties of hair fibers~\cite{Paris2004CaptureOH}, which can lead to failures on certain data sources, such as colorless scans.
An illustration is provided in Fig.~\ref{fig:ablation_filters}.
Thus, unlike previous hair modeling approaches, our method leverages alternative supervision signals -- specifically, denoised 2D edges~\cite{soria2023tiny} and 3D crest lines~\cite{yoshizawa2005fast}. Together, these complementary sources of supervision enable us to achieve superior performance compared to traditional 2D Gabor filters and their 3D counterparts.

\noindent\textbf{Strand-based Hair Priors.}
Hair strands exhibit complex internal structure, making priors essential for accurate modeling. Such priors are typically learned from synthetic datasets, either through generative hairstyle models~\cite{neuralstrands, neuralhaircut, haar, perm, groomgen} or strand-based hair growth models~\cite{deepmvshair, MH, neuralhdhair, GroomCap, hairstep}. Hair growth is commonly formulated as conditional strand generation given a 3D directional field, implemented either via deterministic traversal of the field~\cite{Paris2004CaptureOH} or through neural network–based prediction~\cite{neuralhdhair, MH, GroomCap}.

However, these approaches cannot be used end-to-end during strand-based hairstyle reconstruction and instead depend heavily on the quality of an upstream volumetric hair reconstruction. In contrast, generative hairstyle priors are more flexible and support a wider range of data modalities. They leverage neural generative models, such as latent diffusion~\cite{neuralhaircut, haar}, VAEs~\cite{groomgen}, and GANs~\cite{perm}, and, once trained on synthetic data, can be applied to tasks including reconstruction~\cite{neuralstrands, neuralhaircut, perm}, text-conditioned asset generation~\cite{haar}, and semantic editing~\cite{groomgen, neuralstrands}.
In this work, we employ HAAR~\cite{haar}, a state-of-the-art text-conditional generative prior, to regularize the internal structure of hair during end-to-end reconstruction.

\noindent\textbf{Strand-based Hair Reconstruction.}
The data domains used for strand-based hair reconstruction are diverse and include single images~\cite{hairnet, neuralhdhair, hairstep, Sklyarova2025Im2haircut, difflocks2025}, monocular videos~\cite{neuralhaircut, GH, luo2024gaussianhair, GroomCap, MH}, multi-view captures~\cite{neuralstrands, drhair}, RGB-D input~\cite{zhang2018modeling}, text descriptions~\cite{haar, StrandHead}, and even CT scans~\cite{ct2hair}.
Image-to-hairstyle regression methods train deep networks to directly predict strand-based 3D geometry from 2D cues such as Gabor orientation or depth maps~\cite{hairnet, neuralhdhair, hairstep, Sklyarova2025Im2haircut, perm}. While these methods support convenient in-the-wild usage (e.g., generating assets from a single image), they remain less flexible than generative priors and rely heavily on high-frequency feature extraction to infer strand detail.
Monocular video and multi-view data provide stronger geometric constraints but ultimately suffer from the same reliance on high-frequency observations. RGB-D reconstruction~\cite{zhang2018modeling} fuses front and back views and retrieves the closest hairstyle, but the resulting geometry is still coarse.
Methods that generate hair from text~\cite{haar, StrandHead} cannot preserve personalized or fine-grained hairstyle characteristics.
Hair modeling from CT scans is the closest to our setting. However, CT scanners offer accurate volumetric measurements -- including internal hair structure~\cite{ct2hair} -- which are not available in colorless surface scans.

Our problem setup supports a wide range of applications, from reconstruction using commodity scanners and depth sensors to enriching existing mesh-based hair assets with high-frequency strand geometry.
\section{Method}
\label{sec:method}

In this section, we provide an overview of our method. Our approach is largely inspired by prior works Neural Haircut~\cite{neuralhaircut} and Gaussian Haircut~\cite{GH}. We introduce GeomHair in Subsection~\ref{subsec:GeomHair}, emphasizing the key features of the method, allowing it to be trained directly on 3D scans.
The overview of our method is demonstrated in Fig.~\ref{fig:method}.

\subsection{GeomHair}
\label{subsec:GeomHair}
Our method reconstructs 3D hair strands from full head scans. We begin by segmenting the hair region~\cite{xie2021segformer} and estimating strand orientations by combining geometric crest-line features~\cite{yoshizawa2005fast} with 2D cues extracted from multi-view renderings~\cite{soria2023tiny}. Using the parametric strand generator from~\cite{neuralstrands}, which operates on a latent geometry texture, we synthesize strands that remain surface-constrained and exhibit dense coverage with orientation alignment, following the strategy of~\cite{neuralhaircut}. We further improve reconstruction fidelity by integrating a conditional diffusion prior~\cite{haar} and introducing a multi-step denoising schedule tailored for high-quality strand synthesis.
%



\subsubsection{Estimating Hair Orientation}

\noindent \textbf{3D Crest Lines.} To extract 3D hair strands from high-quality scans, we leverage crest lines -- salient surface features defined by curvature extrema that effectively capture the sharp bends and curves characteristic of hair. Following \cite{yoshizawa2005fast}, crest lines are defined as points on a surface where a principal curvature reaches an extremum along its corresponding direction.


We detect crest lines on 3D hair scans using local cubic polynomial fitting at each mesh vertex. Crest lines are identified where extremality coefficients vanish and satisfy the corresponding convexity/concavity conditions. We retain only the most prominent features by filtering based on line strength and geometric saliency. Each crest line serves as a guide strand for computing local 3D orientations: for each vertex along a crest line, we compute a local coordinate frame via PCA over its neighborhood and refine these frames with adaptive smoothing. The resulting normalized orientations provide supervision for subsequent strand-based reconstruction. Further derivations and implementation details are provided in the supplementary material.

\noindent \textbf{2D orientations.} 
Previous image-based hair strand reconstruction often relies on Gabor filters~\cite{Paris2004CaptureOH} to obtain 2D orientation maps from RGB images, capturing high-frequency hair directionality. However, this approach is not directly applicable to 3D scans, where the lack of color information makes fine-grained orientation inference challenging. 

Specifically, we generate 2D orientation maps by applying the TEED feature extractor~\cite{soria2023tiny} to multi-view renderings of the colorless scans.
During rendering, we employ a forward-facing point light, which we found most effective at highlighting fine mesh details.
After edge detection in the rendered views with TEED, we perform skeletonization to reduce edges to single-pixel paths, effectively suppressing noise and eliminating parasitic lines while preserving the overall hair structure.
The resulting skeletonized paths are converted into a graph representation, from which we extract the longest continuous paths to capture the primary flow of the hair strands.
For each path segment, orientation angles are computed by analyzing the directional vectors between consecutive points. Finally, both the graph structure and its associated orientation vectors are lifted back into 3D space, forming a directional field that supervises the subsequent reconstruction of hair strands.

\begin{figure*}[t!]
    \vspace{-0.2cm}
    \centering
    \includegraphics[width=1\textwidth,trim={0cm 18cm 0cm 0cm},clip]{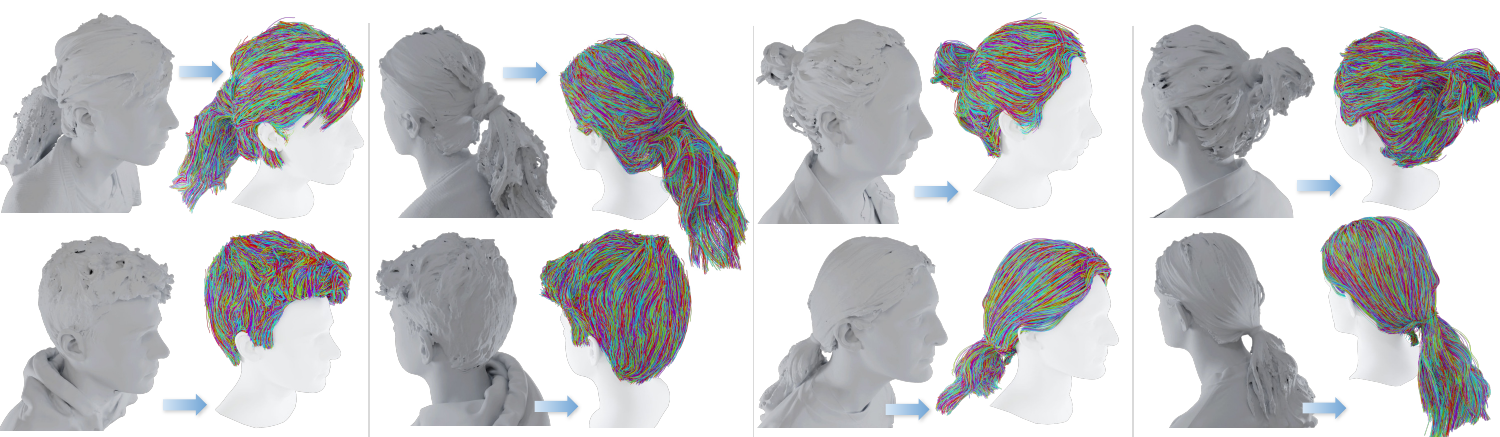}
    \vspace{-0.6cm}
    \captionof{figure}{
    \textbf{Sample scans from Strands400 dataset and the corresponding reconstructions by GeomHair.} 
    }
    \vspace{-0.2cm}
    \label{fig:strands400_inp_vs_out}
\end{figure*}

\subsubsection{Strands-based Reconstruction}
In our framework, a hairstyle is parametrized using a \textit{hair map} $H$ with a resolution of $256\times256$, corresponding to a scalp region of the 3D head model.
Each texel of the map defines a single hair strand that is represented as a 3D polyline with L points: $\S_{i}=\{\p_{i}^{l}\}_{l=1}^{L}$.
%

Our strand-based reconstruction approach is largely inspired by the fine reconstruction stage of Neural Haircut~\cite{neuralhaircut}.
We optimize a latent hairstyle map $\Zgeom = \{\zgeom_{i}\}_{i=1}^{N}$ to match the underlying geometry and orientations, utilizing various supervision signals. 
Here, $N$ is the total number of texels in the hair map, and each latent code $\zgeom_{i}$ is mapped to a single strand $\S_{i}$ by the parametric strand decoder of~\cite{neuralstrands}.


\noindent \textbf{Orientation loss.} Our orientation supervision combines signals from two complementary sources: 3D crest lines and 2D features from the TEED extractor, which are lifted to 3D.
We define the base orientation loss for a given orientation field $\beta$ as:

\begin{equation}\label{eq:l_orient}
\mathcal{L}_\text{orient}(\beta) = \sum_{m=1}^{M} \big(\, 1 - \big| \mathbf{b}_m \cdot \beta(\p_m) \big|\, \big), 
\end{equation}

\noindent where $\mathbf{b}_m$ represents the ground truth orientation at point $\p_m$. The combined orientation loss is then formulated as:

\begin{equation}
\mathcal{L}_\text{orient} = \alpha \cdot \mathcal{L}_\text{orient}(\beta_\text{3D}) + (1-\alpha) \cdot \mathcal{L}_\text{orient}(\beta_\text{2D}),
\label{eq:orient_combined}
\end{equation}

\noindent where $\alpha$ is a weighting parameter, $\beta_\text{3D}$ is the orientation field derived from 3D crest lines, and $\beta_\text{2D}$ is the orientation field from lifted TEED features.
Based on empirical results, we find that $\alpha = 0.5$ provides the best reconstruction quality, creating an equal balance between the 3D and 2D orientation signals.
For each generated strand point near the surface, we apply this orientation loss by finding its nearest neighbor in the supervision signal.

\noindent \textbf{Volume loss.} In prior art~\cite{neuralhaircut}, signed distance functions (SDFs) are used as a continuous representation of a head volume, as well as volume between the head surface and hair, bordering with the outside space. 
In our method, we utilize an unsigned distance function (UDF) approximator~\cite{zhou2023levelset}.
This choice is motivated by the fact that our hair mesh, extracted from scans, may not be watertight, thus lacking internal structure, making the use of UDF more suitable for our scenario.
In particular, we pretrain the UDF model $F_{\text{udf}}: \mathbb{R}^3 \rightarrow \mathbb{R}^+$ on our extracted hair mesh.
During optimization, we use the extracted UDF to supervise the strands with the volume loss:

\begin{equation}
    \mathcal{L}_\text{volume} = \sum_{i=1}^{N} \sum_{l=1}^L \big( F_{\text{udf}}(\p^l_i) \big)^2,
    \label{eq:vol_udf}
\end{equation}

\noindent where $\p^l_i$ represents the $l$-th point of the $i$-th strand.

\noindent \textbf{Coverage loss.} To ensure uniform coverage of the visible outer surface of the scan with strads, we apply a one-way Chamfer loss $\mathcal{L}_\text{chamfer}$, following~\cite{neuralhaircut}:

\begin{equation}
    \mathcal{L}_\text{chamfer} = \sum_{q=1}^Q \big\| \x_q - \p_q \big\|_2^2,
\end{equation}

\noindent where $\x_q$ are sampled points on the scan and $\p_q$ are their nearest neighbors among the strands set.



\noindent \textbf{Diffusion prior.}
%
Diffusion-based loss $\mathcal{L}_\text{prior}$ is a crucial component to enhance realism of the optimized hairstyle~\cite{neuralhaircut, GH} by forcing $\Zgeom$ to be in the distribution of latent maps of only plausible hairstyles. 

We adopt the conditional diffusion model from HAAR~\cite{haar}, applied to $\Zgeom_\text{LR}$, the low-resolution version of $\Zgeom$ matching the HAAR diffusion input resolution. Hairstyle descriptions are generated by a VQA model~\cite{liu2023llava} from scan-shading renderings and processed through BLIP~\cite{li2022blip} into a text embedding $\tau(P)$ that conditions the diffusion model.

We refer to $\sigma_i$ as the \emph{noise level} at optimization iteration $i$ -- the standard deviation of the Gaussian perturbation in $\hat{\Zgeom}_\text{LR} = \Zgeom_\text{LR} + \epsilon \cdot \sigma_i$, with $\epsilon \sim \mathcal{N}(0, I)$, following the EDM formulation~\cite{Karras2022ElucidatingTD}. A denoiser network $\mathcal{D}_\theta$ estimates the clean latent:
\begin{equation}
    \mathcal{D}_\theta (\hat{\Zgeom}_\text{LR}, \sigma_i, P) = c_\text{skip} \hat{\Zgeom}_\text{LR} + c_\text{out} \mathcal{F}_\theta \big( c_\text{in} \hat{\Zgeom}_\text{LR},\, c_\text{noise},\, \tau (P) \big),
\end{equation}

\noindent where $c_\text{skip}, c_\text{out}, c_\text{in}, c_\text{noise}$ are the standard EDM preconditioning coefficients (deterministic functions of $\sigma_i$) and $\mathcal{F}_\theta$ is the HAAR pre-trained U-Net, kept \emph{frozen} during optimization.

We use a two-phase noise schedule based on the Karras discretization~\cite{Karras2022ElucidatingTD}. For the first $N_\text{warmup}$ iterations, $\sigma_i$ follows a precomputed $N_\text{warmup}$-element Karras schedule that monotonically decreases from $\sigma_\text{max}$ to $\sigma_\text{min}$, and at each iteration we additionally run Euler-ancestral sampling along a local $K$-element schedule from $\sigma_i$ down to $\sigma_\text{min}$. Afterwards, $\sigma_i$ is sampled from a fixed cosine-interpolated density over $[\sigma_\text{min}, \sigma_\text{max}]$ and a single denoiser pass is applied.

The diffusion prior loss in both phases is:
\begin{equation}
    \mathcal{L}_\textrm{prior} =  \mathbb{E}_{\sigma_i, \epsilon, \Zgeom_\text{LR}, P} \big[ \lambda_i \| \mathcal{D}_\theta (\hat{\Zgeom}_\text{LR}, \sigma_i, P) - \Zgeom_\text{LR} \|_2^2 \big],
\end{equation}

\noindent where $\lambda_i$ is a noise-level-dependent weight.
%


\noindent \textbf{Overall objective.} 
Our final optimization objective combines geometric and diffusion-based constraints:

\vspace{-0.4cm}
\begin{equation*}
    \mathcal{L}_\text{final} = \mathcal{L}_\text{volume}\, +\, \lambda_\text{chamfer} \mathcal{L}_\text{chamfer}\, +\, \lambda_\text{orient} \mathcal{L}_\text{orient}\, +\, \lambda_\text{prior} \mathcal{L}_\text{prior}
    \label{eq:final}
\end{equation*}
\section{Experiments}
\label{sec:experiments}

We collect an extended version of the NPHM~\cite{giebenhain2023learning} dataset that consists of scans captured by Artec Eva scanners. We fit FLAME~\cite{flame} head to each scan and segment the hair region by applying \cite{xie2021segformer}.

\subsection{Implementation details}
\label{subsec:setup}

We train our model on a single GPU RTX A6000 for 75,000 iterations using the Adam optimizer, which employs a learning rate of 0.001 and a step-wise multi-step scheduler with a gamma value of 0.5. The whole training typically takes about 14 hours.
For the diffusion prior, we set $\sigma_\text{max} = 80$ and $\sigma_\text{min} = 0.5$. The warmup phase covers $N_\text{warmup} = 10{,}000$ iterations, during which we run $K = 2$ Euler-ancestral denoising steps~\cite{Karras2022ElucidatingTD} per iteration with a classifier-free guidance weight of $4.0$; afterwards, we sample $\sigma_i$ from the cosine-interpolated density and use a single denoiser pass.

\begin{figure*}[t!]
    \centering
    \includegraphics[width=1.0\linewidth]{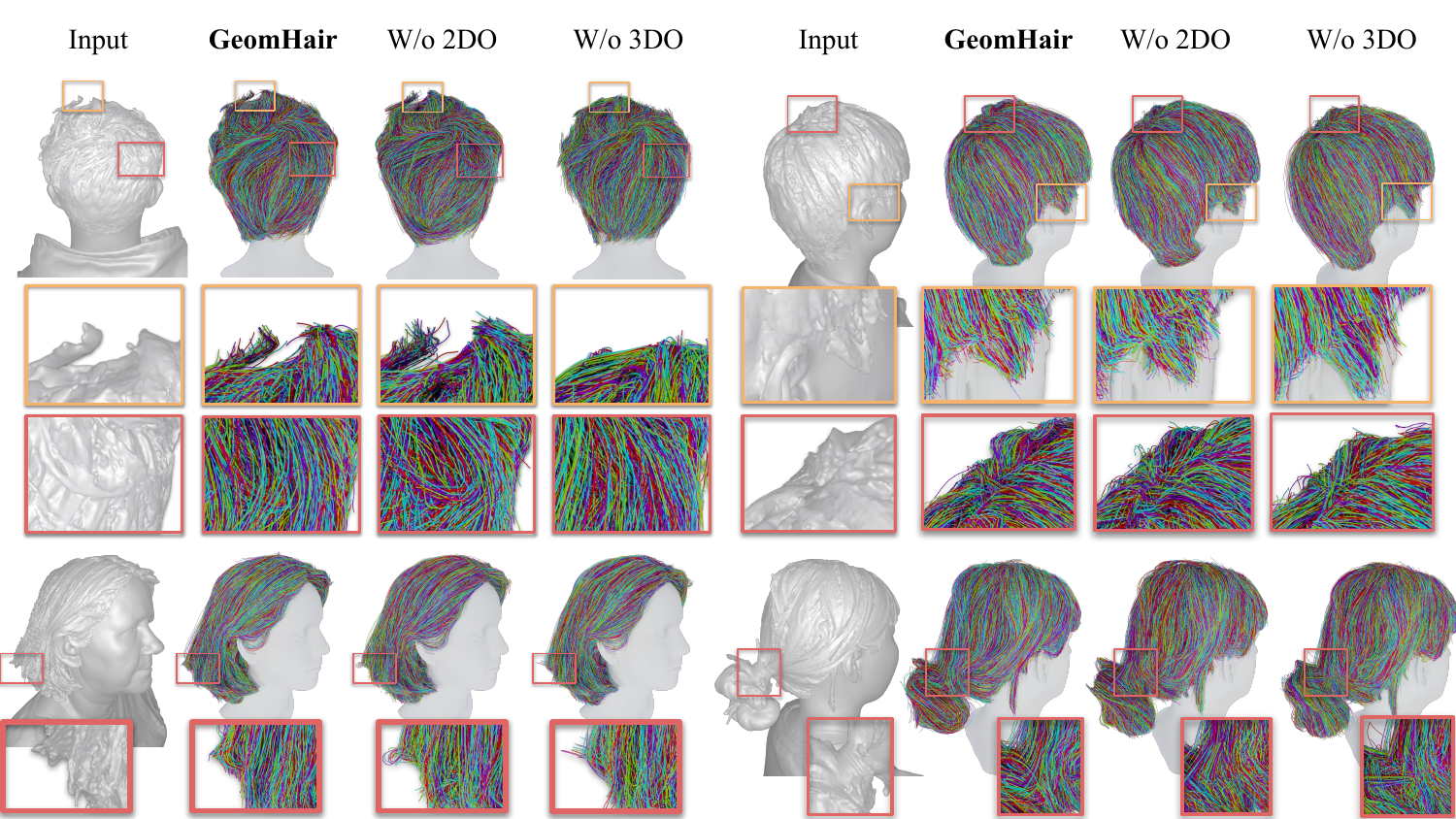}
    \vspace{-0.6cm}
    \caption{\textbf{Qualitative ablation of orientation sources.} Combining 3D crest lines and 2D TEED orientations (Ours) produces more coherent and detailed strands than using either source alone.}
    \label{fig:orientation_ablation}
    \vspace{-0.2cm}
\end{figure*}

\subsection{Results}
\label{subsec:results}

\begin{figure*}[t]
    \centering
    \includegraphics[width=\linewidth]{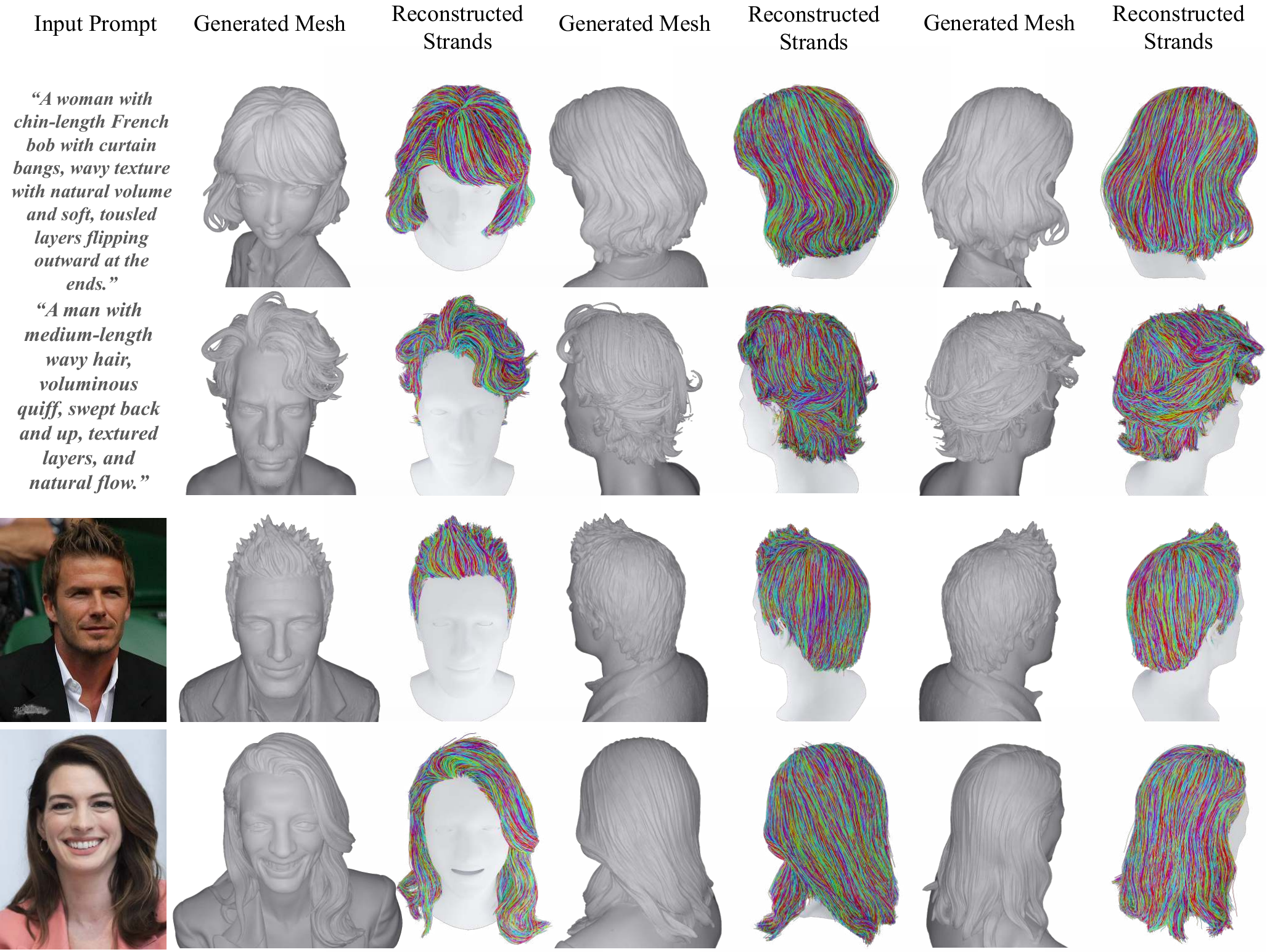}
    \vspace{-0.4cm}
    \caption{
    \textbf{Text-to-Strands and Single image-to-Strands.}
    Our method can be used to generate hair strands from an input text or image prompt by leveraging off-the-shelf mesh generation methods as an intermediary step.
    Here, we relied on~\cite{meshy2025} as an off-the-shelf text-to-mesh / image-to-mesh proxy. 
    }
    \vspace{-0.2cm}
    \label{fig:strands_from_meshy}
\end{figure*}
\begin{figure*}[t]
    \centering
    \includegraphics[width=\linewidth]{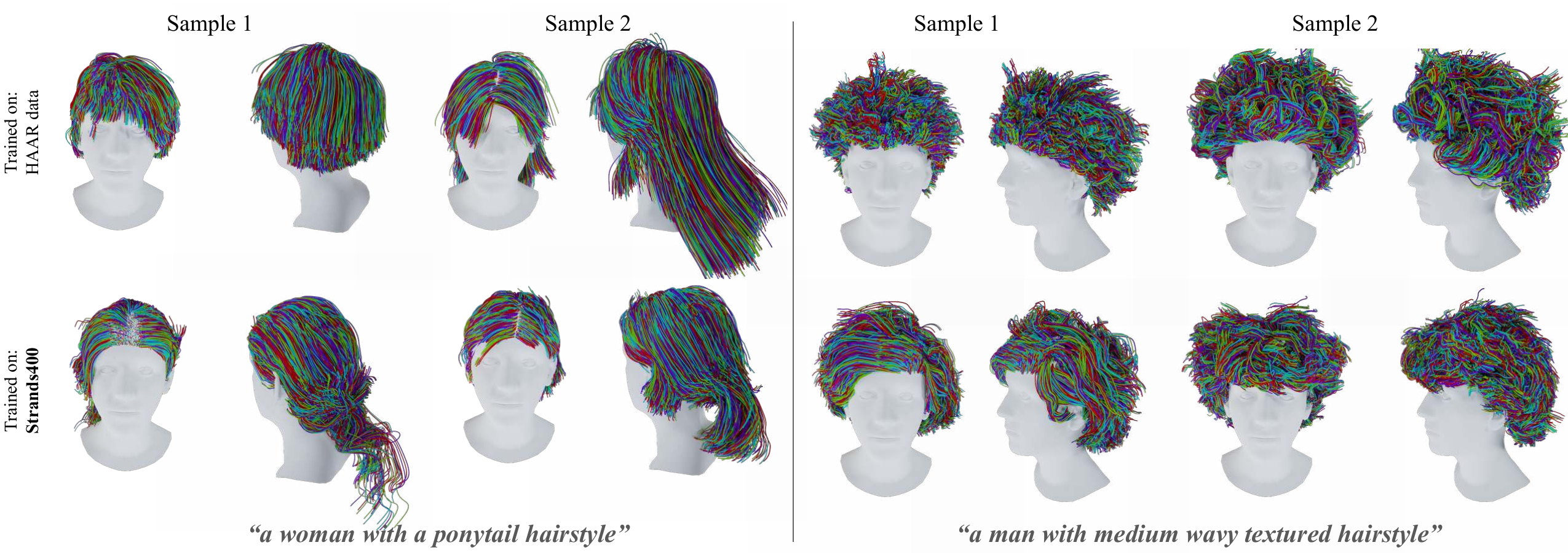}
    \vspace{-0.4cm}
    \caption{\textbf{Qualitative comparison of HAAR trained on different datasets}. Top row shows results when trained on the original synthetic HAAR dataset, while bottom row shows results when trained on Strands400 dataset. Both models generate strands from the same prompts.}
    \label{fig:haar_comparison}
    \vspace{-0.3cm}
\end{figure*}

\begin{figure}[t!]
    \centering
    \includegraphics[trim={0 0 0 0},clip,width=0.4\textwidth]{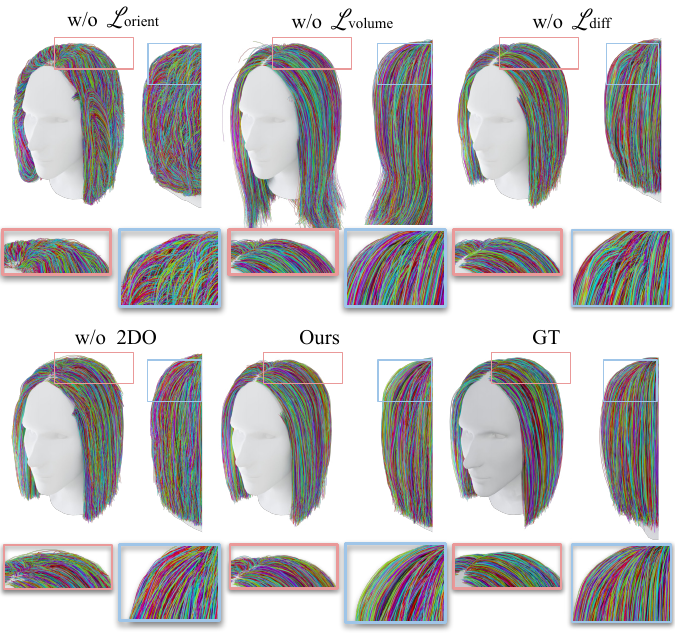}
    \vspace{-0.3cm}
    \caption{\textbf{Ablation on a straight-hair sample.}
    We ablate the most important components of the GeomHair pipeline on an artist-drawn straight-hair sample (\textit{wStraight.hair}) from Cem Yuksel's Hair Models~\cite{cem_yuksel}.
    }
    \label{fig:ablation}
    \vspace{-0.2cm}
\end{figure}

\begin{figure}[h!]
    \centering
    \includegraphics[width=1.0\linewidth]{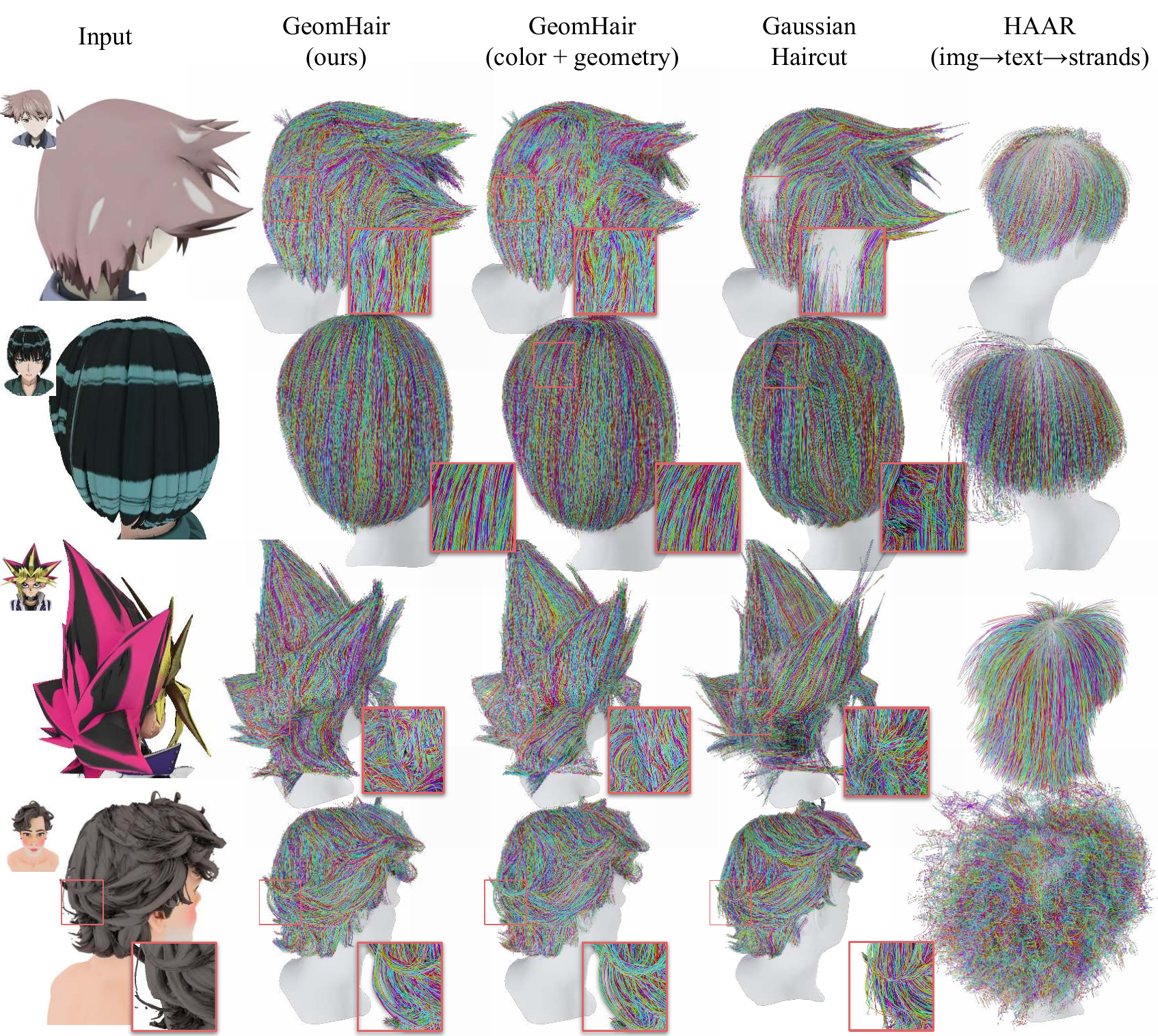}
    \vspace{-0.6cm}
    \caption{\textbf{Comparison of supervision signals on generated assets with RGB texture.} We compare geometric supervision (Ours), combined geometric+RGB supervision, RGB-only supervision (Gaussian Haircut~\cite{GH}), and HAAR~\cite{haar}. Geometric supervision alone produces high-quality results; adding RGB provides marginal benefit, while removing geometry degrades quality.}
    \label{fig:supervision_comparison}
    \vspace{-0.4cm}
\end{figure}

\noindent \textbf{Qualitative evaluation.} We evaluate the effect of different supervision signals on reconstruction quality using generated assets with available RGB texture. As shown in Fig.~\ref{fig:supervision_comparison}, we compare geometric supervision alone (our method), combined geometric and RGB supervision, RGB-only supervision (Gaussian Haircut~\cite{GH}), and the HAAR~\cite{haar} generative baseline. Adding RGB on top of geometry yields marginal gains, while relying solely on RGB noticeably harms the results. HAAR, which generates strands without per-scene optimization, produces less faithful reconstructions, demonstrating the sufficiency of our geometry-based approach.

\begin{figure}[h!]
    \centering
    \vspace{-0.3cm}  
    \includegraphics[trim={0cm 1.2cm 0cm 0cm},clip,width=0.45\textwidth]{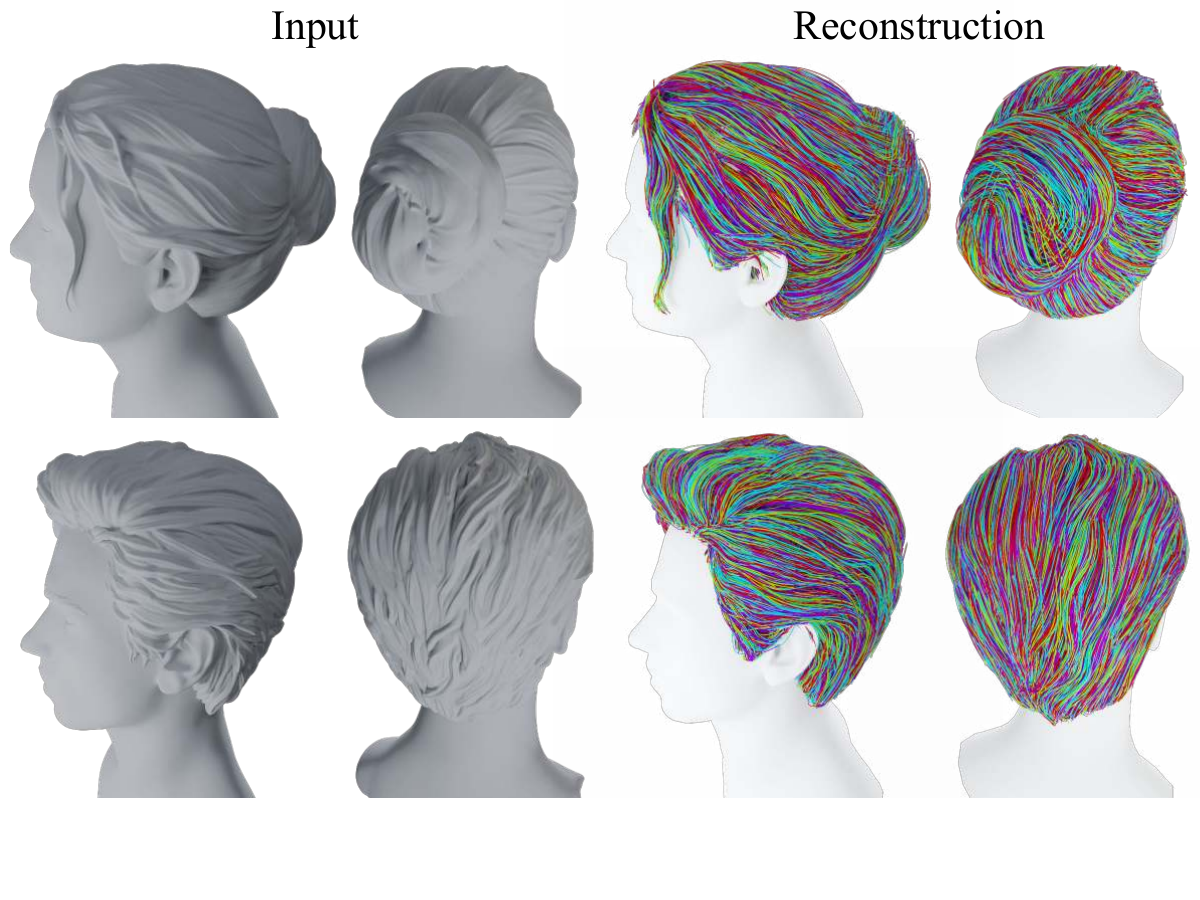}
    \vspace{-0.3cm}  
    \caption{\textbf{Results on 3D-designed assets.}
    Here, we demonstrate the results of fitting to a synthetic, hand-carved 3D mesh from a 3D stock (\textit{CGTrader, author: ZStuff}).
    Our method is capable of reconstructing a plausible collection of strands even when fewer guiding strands can be observed.
    }
    \label{fig:blender_asset}
     \vspace{-0.3cm}
\end{figure}

\noindent \textbf{Ablation study.} We conduct a comprehensive ablation study highlighting each component's significance, see Fig.~\ref{fig:ablation}. Removing orientation loss ($\mathcal{L}_\text{orient}$) produces unrealistic directions, excluding volume loss ($\mathcal{L}_\text{volume}$) extends strands beyond plausible regions, and omitting diffusion prior ($\mathcal{L}_\text{prior}$) reduces structural realism.

Next, we ablate the contributions of 3D crest lines and 2D TEED-based orientations, denoted as ``w/o 3DO'' and ``w/o 2DO'', respectively. Removing crest lines severely impacts the model’s ability to infer plausible hair structure. 
Interestingly, omitting 2D orientations slightly improves quantitative metrics but degrades visual quality (Fig.~\ref{fig:orientation_ablation}): only 3D crest lines (w/o 2DO) produces overly smooth strands lacking fine directional variation, while only 2D orientations (w/o 3DO) yields noisier results without coherent global structure. Combining both (Ours) strikes the best balance, motivating this choice despite the marginal quantitative trade-off.

We also compare 2D Gabor filters, the TEED neural extractor, and Crest Lines in Fig.~\ref{fig:ablation_filters}. Following~\cite{neuralhaircut}, we filter the Gabor orientation map by estimated confidence, but observe noisier results than other extractors and sensitivity to many hyperparameters.

\begin{figure}[h!]
    \centering
    \includegraphics[width=1.0\linewidth]{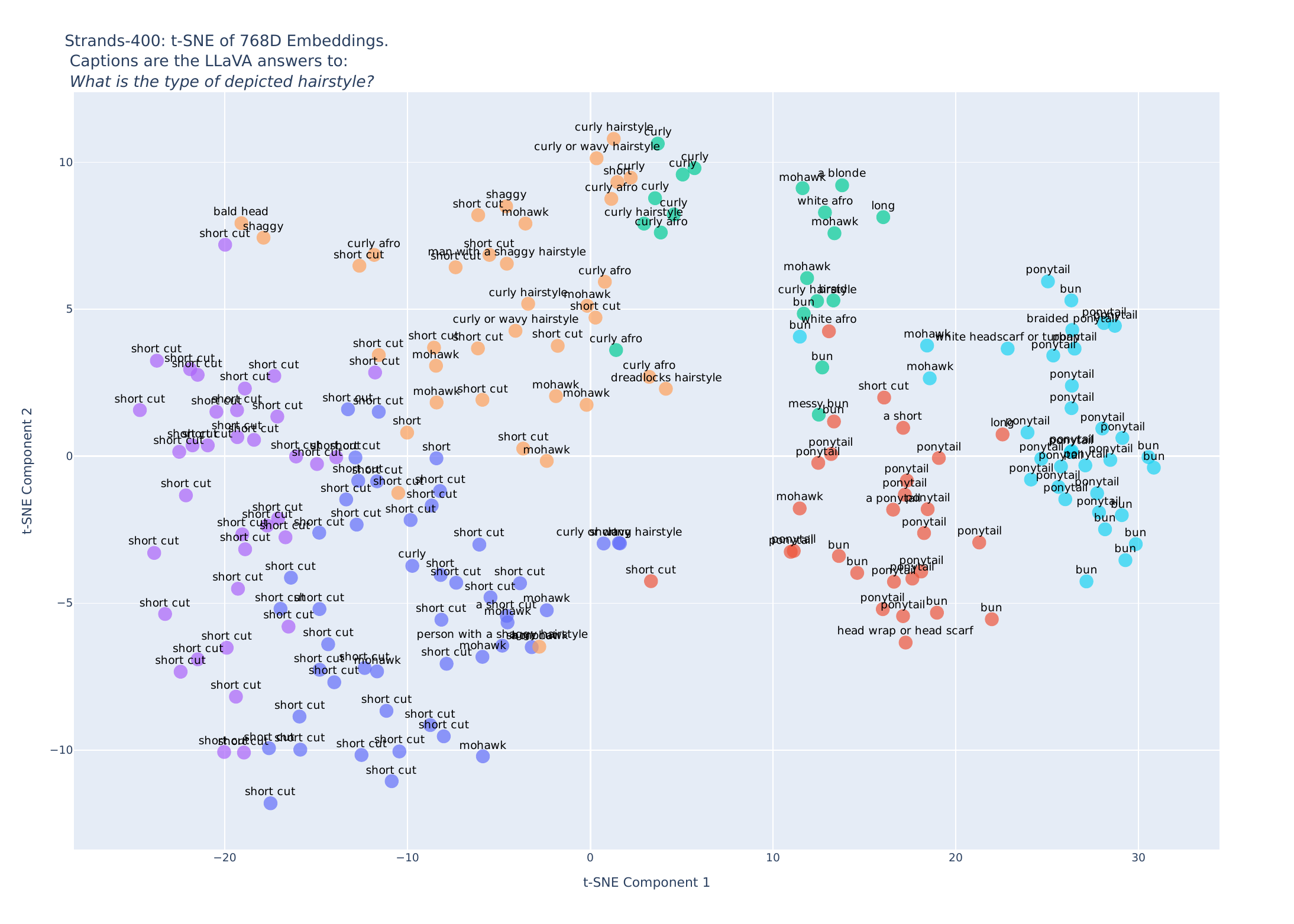}
    \vspace{-0.6cm}
    \caption{\textbf{Hairstyle type distribution in Strands400.} t-SNE~\cite{van2008visualizing} over BLIP embeddings~\cite{li2022blip} of LLaVA~\cite{liu2023llava} answers, colored by K-Means~\cite{ahmed2020k} clusters, reveals clear groupings by hairstyle type (e.g.\ short cut, curly, ponytail, bun).}
    \label{fig:tsne_length}
    \vspace{-0.4cm}
\end{figure}

\noindent \textbf{Strands400 dataset.}
Based on the reconstructions of GeomHair, we compile a dataset of 400 different people with corresponding strands.
The dataset consists of 383 manually curated samples from the NPHM dataset~\cite{giebenhain2023learning} that yield the best reconstructions and 17 newly collected samples captured with a similar dual Artec Eva scanner setup.
We demonstrate the scans and their reconstructions from the Strands400 dataset in Fig.~\ref{fig:strands400_inp_vs_out}. 
To assess realism, we conducted a perceptual evaluation comparing rendered hairstyles from Strands400 and the Perm~\cite{perm} training set using GPT-4V-based side-by-side preference judgments. Strands400 was preferred in 74.4\% of cases.
Additionally, we analyze the content of the answers provided by the VQA model (LLaVA~\cite{liu2023llava}) after showing the rendered shading of the frontal and back views of the 3D scans. We use the same BLIP embedder~\cite{li2022blip} as in training to obtain embeddings from the answers to all 27 questions and fit t-SNE~\cite{van2008visualizing} to visualize them on a 2D plane. To obtain the coloring of the t-SNE points, we run K-Means~\cite{ahmed2020k} over 5 clusters. As shown in Fig.~\ref{fig:tsne_length}, the resulting clusters reveal clear semantic groupings by hairstyle type (e.g.\ short cut, curly, ponytail, bun), confirming that the dataset spans a diverse and well-separated range of hairstyles.

\noindent \textbf{Applications.} We show the potential of our approach for dataset creation from mesh-based hair assets. We demonstrate results on artist-created hairstyles (Fig.~\ref{fig:blender_asset}) and on hair strands generated from text and image prompts via off-the-shelf mesh generation~\cite{meshy2025} (Fig.~\ref{fig:strands_from_meshy}), purely from geometry without color information. Extended text-to-strands and image-to-strands results are provided in the supplementary material. Lastly, we demonstrate the viability of the Strands400 dataset by retraining the HAAR generative model on our dataset instead of the original synthetic training data. As shown in Fig.~\ref{fig:haar_comparison}, the model generates more realistic ponytails and wavy hairstyles.

\vspace{-0.2cm}
\section{Conclusion}
\label{sec:conclusion}
We propose a method for reconstructing human hair directly from colorless 3D geometry by combining geometric orientation cues with a diffusion prior to guide strand generation from the surface. Qualitative evaluations and ablations on collected scans, designer assets, and generated meshes show that our method reconstructs challenging hairstyles from geometry alone. We further leverage GeomHair to compile Strands400, the largest publicly available real-world strand dataset, enabling training of generative models for downstream tasks. 

\nocite{abdi2010principal}
{
    \small
    \bibliographystyle{ieeenat_fullname}
    \bibliography{main}
}


\pagebreak

\appendix
\section{Technical Details}
\label{sec:supp_technical_details}

\noindent \textbf{Questions employed in the calculation of the text prompt for the diffusion prior conditioning}.
We provide the list of questions for querying LLaVA model below.

\begin{enumerate}
    \item \textit{Describe in detail the bang/fringe of depicted hairstyle including its directionality, texture and coverage of face?}
    \item \textit{What is the overall hairstyle depicted in the image? }
    \item \textit{Does the depicted hairstyle longer than the shoulders or shorter than the shoulder?}
    \item \textit{Does the depicted hairstyle has short bang or long bang or no bang from frontal view? }
    \item \textit{Does the hairstyle has straight bang or Baby Bangs or Arched Bangs or Asymmetrical Bangs or Pin-Up Bangs or Choppy Bangs or curtain bang or side swept bang or no bang?}
    \item \textit{Are there any afro features in the hairstyle or no afro features? }
    \item \textit{Is the length of hairstyle shorter than middle of the neck or longer than middle of the neck?}
    \item \textit{What is the main geometry features of the depicted hairstyle? }
    \item \textit{What is the overall shape of the depicted hairstyle?}
    \item \textit{Is the hair short, medium, or long in terms of length?}
    \item \textit{What is the type of depicted hairstyle}
    \item \textit{What is the length of hairstyle relative to human body?}
    \item \textit{Describe the texture and pattern of hair in the image.}
    \item \textit{What is the texture of depicted hairstyle}
    \item \textit{Does the depicted hairstyle is straight or wavy or curly or kinky?}
    \item \textit{Can you describe the overall flow and directionality of strands?}
    \item \textit{Could you describe the bang of depicted hairstyle including its directionality and texture}
    \item \textit{Describe the main geometric features of the hairstyle depicted in the image}
    \item \textit{Is the length of hairstyle buzz cut, pixie, ear length, chin length, neck length, shoulder length, armpit length or mid-back length?}
    \item \textit{Describe actors with similar hairstyle type. }
    \item \textit{Does the haistyle cover any parts of the face? Write which exactly parts. }
    \item \textit{In what ways is this hairstyle a blend or combination of other popular hairstyles? }
    \item \textit{Could you provide the most closest types of hairstyles from which this one could be blended? }
    \item \textit{How adaptable is this hairstyle for various occasions (casual, formal, athletic)? }
    \item \textit{How is this hairstyle perceived in different social or professional settings?}
    \item \textit{Are there historical figures who were iconic for wearing this hairstyle?}
    \item \textit{Could you describe the partition of this hairstyle if it is visible?}
\end{enumerate}

~

\noindent\textbf{3D orientation extraction with crest lines.}
Following \cite{yoshizawa2005fast}, we define crest lines on a surface $S$ as the set of points where one of the principal curvatures reaches an extremum along its corresponding curvature direction.
Mathematically, we can express this as:
\begin{equation}
e_\text{max} = \frac{\partial k_\text{max}}{\partial t_\text{max}} = 0 \quad \text{(for convex crest lines)},
\end{equation}
\begin{equation}
e_\text{min} = \frac{\partial k_\text{min}}{\partial t_\text{min}} = 0 \quad \text{(for concave crest lines)},
\end{equation}
where $k_\text{max}$ and $k_\text{min}$ are the maximum and minimum principal curvatures, with corresponding principal directions $t_\text{max}$ and $t_\text{min}$. The associated extremality coefficients are $e_\text{max}$ and $e_\text{min}$. 

We identify crest lines on 3D hair scans using local cubic polynomial fitting at each mesh vertex, of the form:
\begin{align*}
h(x, y) = &\frac{1}{2}(b_0x^2 + 2b_1xy + b_2y^2) + \\
          &\frac{1}{6}(d_0x^3 + 3d_1x^2y + 3d_2xy^2 + d_3y^3)
\end{align*}
After computing these values, crest lines are traced across the mesh by connecting points where the extremality coefficients vanish. To retain only the most salient hair features, we apply thresholding based on the cyclideness measure:
\begin{equation}
C = \sqrt{|e_\text{max}|^2 + |e_\text{min}|^2}
\end{equation}
Let $C = {\{c_1, \ldots, c_N\}}$ be a set of crest lines, where each $c_i=\{\p_i\}_{i=1}^{L_{i}}$ consists of $L_i$ points, which may vary across crest lines.
Each crest line is treated as a guide hair strand for computing local orientations.
We estimate local curvature along each crest line $c_i$ to guide adaptive window sizing. 
For each point $\p_i$, a local coordinate frame is computed via PCA \cite{abdi2010principal} over its neighborhood. 
An adaptive smoothing step refines these frames, balancing noise reduction and preservation of directional variation. 
The resulting normalized orientations are used as supervision for strand-based reconstruction.

\begin{figure}[t!]
    \vspace{-0.1cm}
    \centering
    \includegraphics[clip,trim={0cm 0cm 0cm 0cm},width=\linewidth]{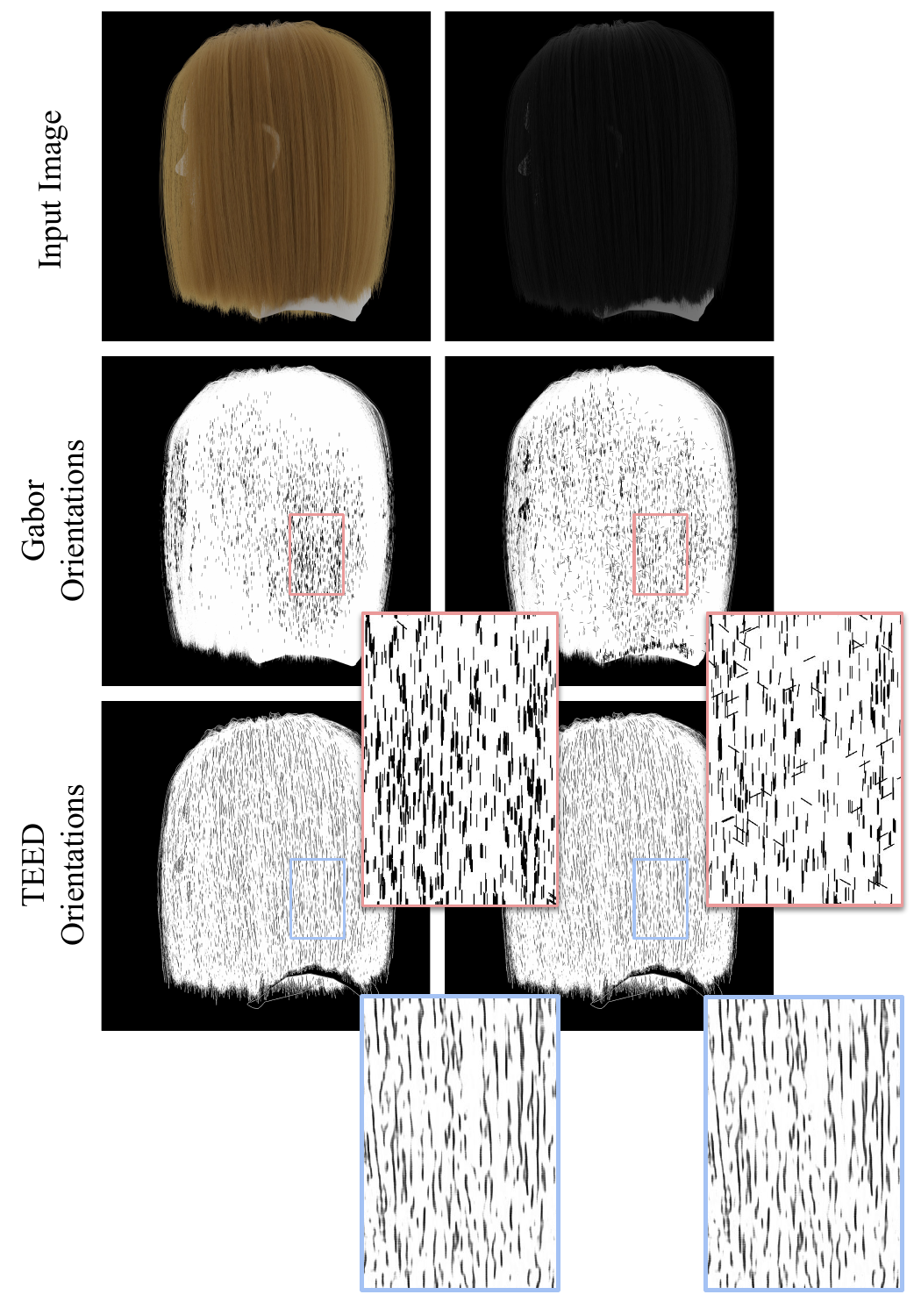}
    \vspace{-0.65cm}
    \caption{
    Commonly used hair orientation detector in RGB -- Gabor filters -- is sensitive to the hair color. 
    The lighter color (left row) yields a better orientation than its darker (right row) counterpart.
    On the other hand, orientation detected from TEED is more robust to hair color.
    This example demonstrates that, even under decent capture conditions, hair reconstruction from multi-view RGB images can highly depend on features like hair color, unlike the typical reconstruction from geometry obtained from a structured-light 3D scanner.
    }
    \label{fig:gabor_color_variance}
    \vspace{-0.4cm}
\end{figure}

\noindent\textbf{Gabor filter sensitivity to hair color.} We demonstrate an experiment where the same hairstyle produces different extracted orientations when Gabor filters are applied to different hair colors. Fig. \ref{fig:gabor_color_variance} shows that the orientation field obtained from lighter hair matches the straight hairstyle more closely, while the orientation field from darker hair appears noticably noisier.
This contrived example demonstrates that, even under decent capture conditions, hair reconstruction from multi-view RGB images can highly depend on features like hair color, unlike the typical reconstruction from geometry obtained from a structured-light 3D scanner.

\section{Strands400 Dataset}

\vspace{0.2cm}
\noindent \textbf{Legal notice.}
All participants in our dataset signed an agreement form
compliant with GDPR. Please note that GDPR compliance
includes the right for every participant to request the timely
deletion of their data, which we will enforce as part of the
distribution process of our dataset.\\

\noindent \textbf{Dataset statistics.}
In Fig.~\ref{fig:dataset_stats}, we demonstrate the distribution of the age, stratified by gender, (left) and of ethnicity (right), reported by the participants in the Strands400 dataset.
In the age histogram (left), the horizontal axis corresponds to the participants' age bins, and the vertical axis corresponds to the number of people in the bin.
Only the votes of the participants who willingly disclosed that information were taken into account. 

\begin{figure*}
    \includegraphics[clip,trim={0cm 0cm 0cm 0cm},width=\linewidth]{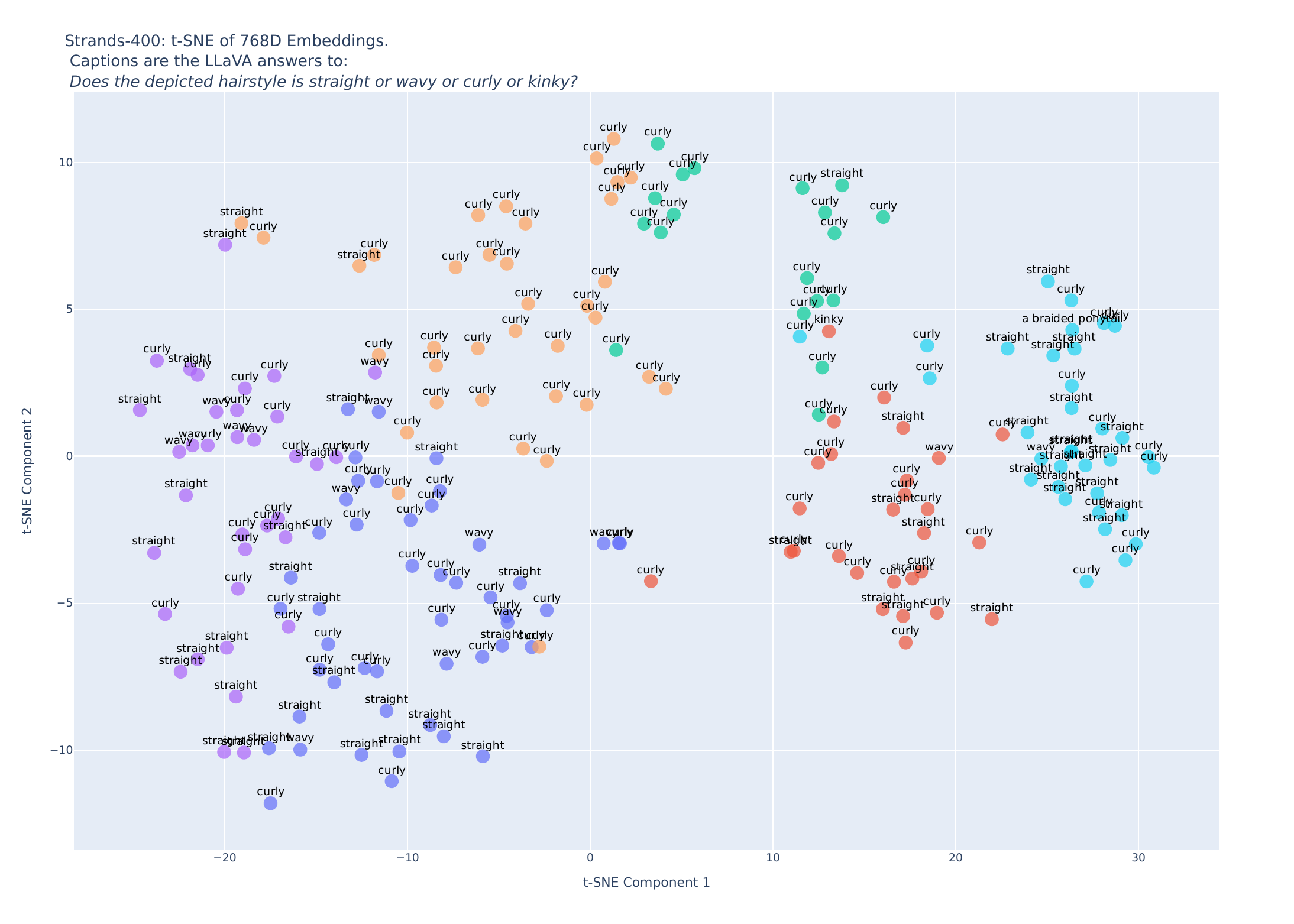}
    \caption{The distribution of hair waviness in the Strands400 dataset.
    The captions are collected from the answers of a VQA model (LLaVA~\cite{liu2023llava}) after showing the rendered shading of the frontal and back views of the 3D scans in Strands400.
    The locations correspond to the t-SNE~\cite{van2008visualizing} over BLIP embeddings~\cite{li2022blip} of the LLaVA answers.
    The colors are calculated via K-Means~\cite{ahmed2020k}.}
    \label{fig:tsne_waviness}
\end{figure*}

In the main paper, we visualize the t-SNE projection of VQA embeddings colored by hairstyle type. Here, we additionally show the projection colored by hair waviness in Fig.~\ref{fig:tsne_waviness}. Using the same BLIP embeddings~\cite{li2022blip} of LLaVA~\cite{liu2023llava} answers and K-Means~\cite{ahmed2020k} clustering with 5 clusters, we observe clear groupings that separate straight, wavy, and curly hairstyles. The t-SNE locations are enhanced with LLaVA answers for the respective samples, and each second sample is shown to provide more space for the captions. Together with the hairstyle type visualization in the main paper, these plots confirm that the Strands400 dataset captures a broad spectrum of hairstyle attributes.\\

\begin{table*}[h!]
    \vspace{-0.1cm}
    {
        \setlength{\tabcolsep}{0pt}
        \renewcommand{\arraystretch}{0}
        \begin{tabular}{ccc}
            \includegraphics[clip,trim={0cm 0cm 0cm 0cm},width=0.6\linewidth]{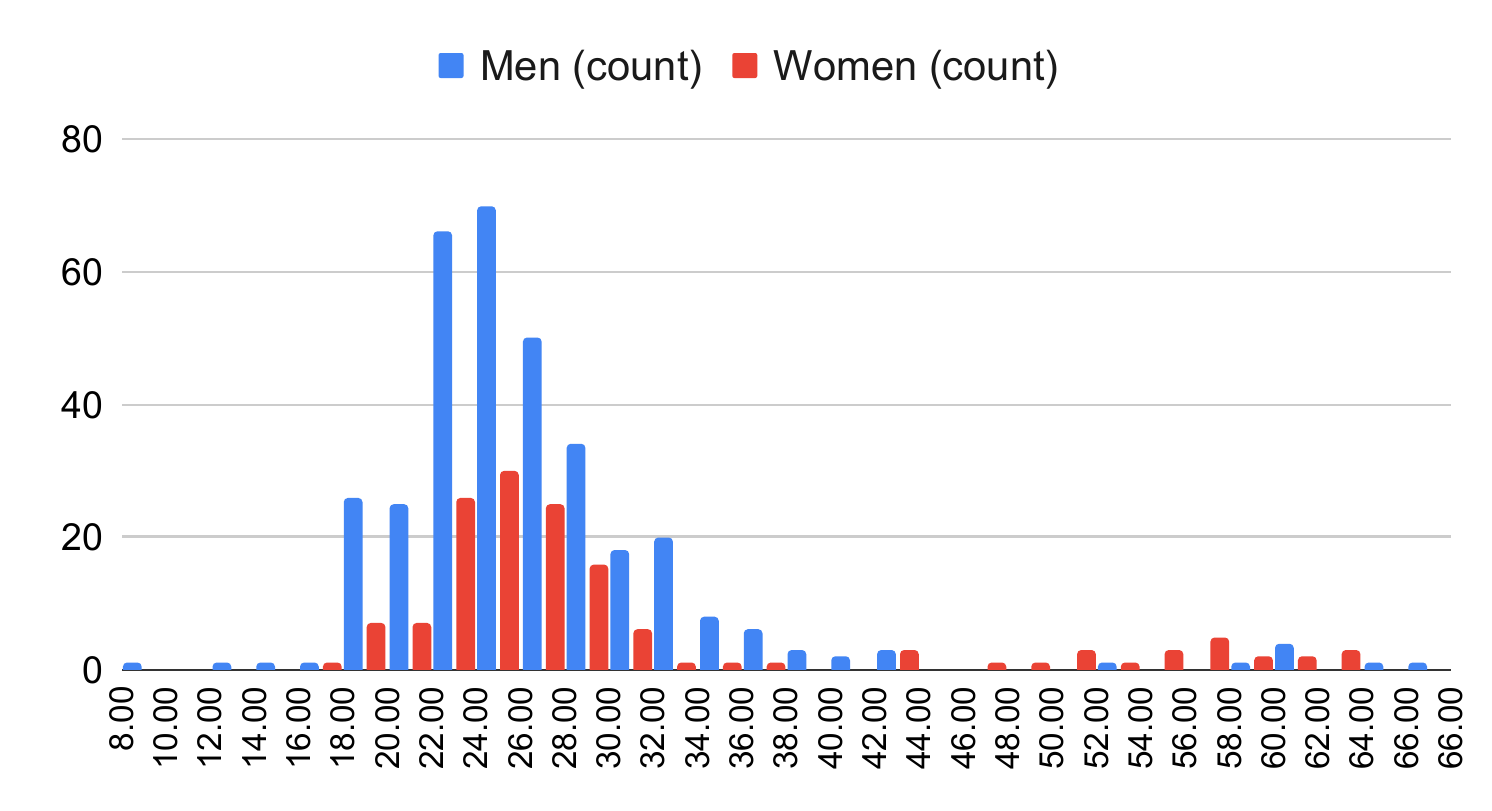}
        &
            \includegraphics[clip,trim={0cm 0cm 0cm 0cm},width=0.4\linewidth]{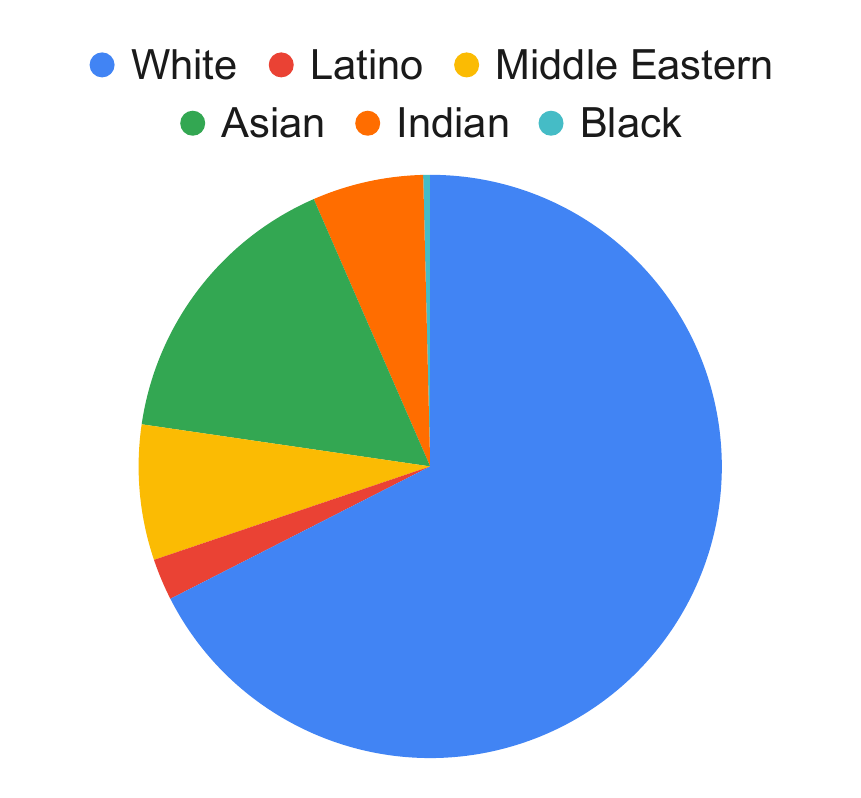}
        \\
        \end{tabular}
    }
    \\[0.2cm]
    \captionof{figure}{
    The distribution of the age, stratified by gender, (left) and of ethnicity (right), reported by the participants in the Strands400 dataset.
    Only the votes of the participants who willingly disclosed that information were taken into account. 
    }
    \label{fig:dataset_stats}
\end{table*}

\noindent \textbf{Capture setup.}
Our dataset consists of two parts -- the subset of the latest version of NPHM that contains 383 scans, and additionally collected 17 scans with a setup similar to the NPHM setup. We processed 472 scans from NPHM dataset and categorize the reconstructed strands into three categories: good, improvable, and poor. This process yielded 322 high-quality samples. Additionally, we enhanced 61 samples from the second category by removing facial hair that had been inadvertently included from hair segmentation errors, originally happened during preprocessing. Facial hair was removed by manually selecting the polygons in 2D, defining the hair regions to exclude. Finally, we collected 24 samples separately and exclude the 7 samples from the dataset.
The additional scans collection setup consists of two handheld Artec Eva scanners, rotating over a $360^\circ$ trajectory within $\sim$ 3 seconds to capture a single person. 
Since our setup largely follows NPHM capture setup, we refer the reader to the NPHM paper for the remaining details regarding the capture setup~\cite{giebenhain2023learning}.
The participants in this category were selected with an emphasis on hairstyles, more challenging for reconstruction (wavy, curly, etc.), to better align the overall distribution to the overall spectrum of hairstyles.\\

\begin{figure*}
    \centering
    \includegraphics[width=1\textwidth,trim={0cm 0cm 0cm 0cm},clip]{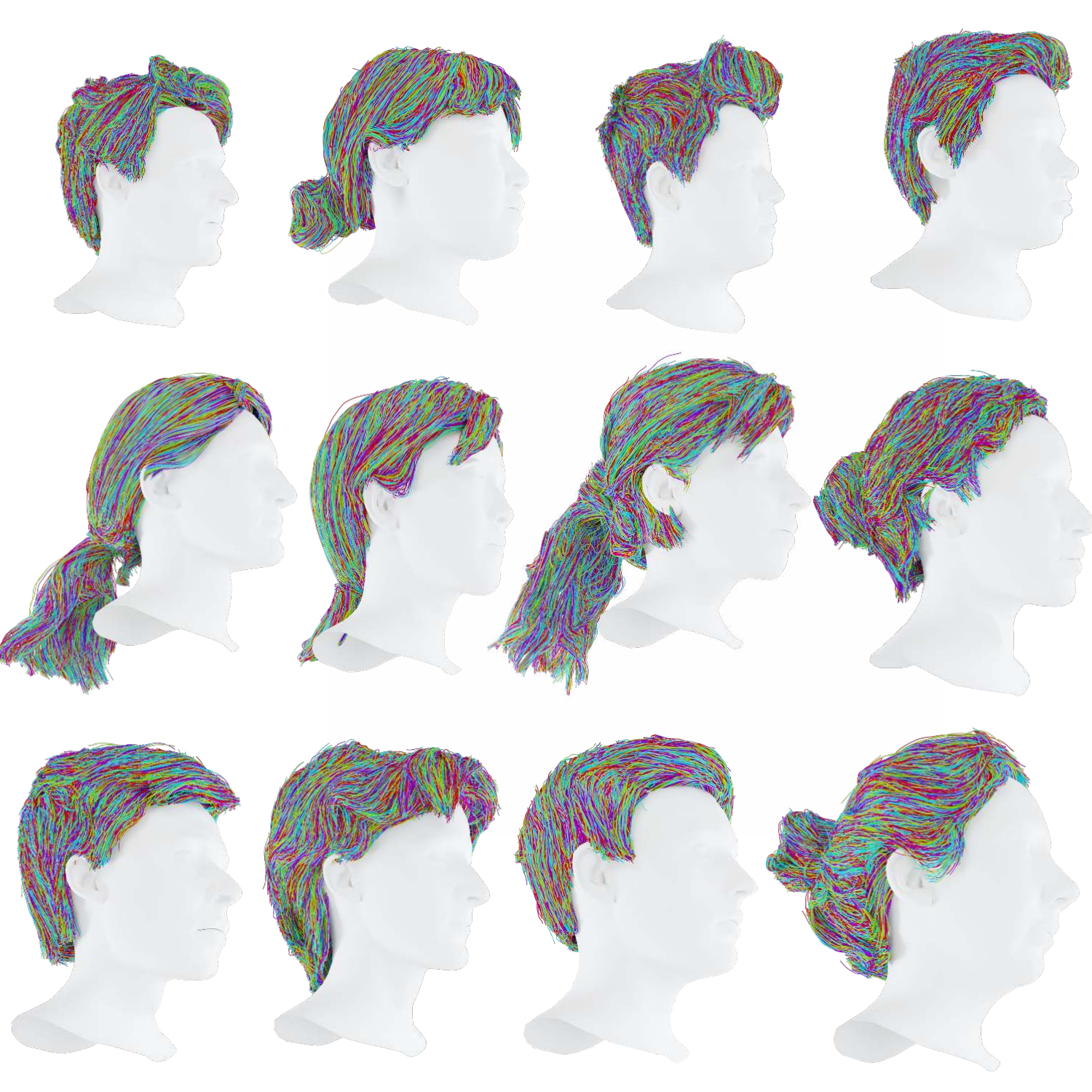}
    \vspace{-0.5cm}
    \captionof{figure}{
    Sample strands reconstructions from Strands400 dataset.
    }
    \vspace{-0.3cm}
    \label{fig:strands400}
\end{figure*}

\noindent \textbf{Representative samples.}
More samples from the Strands400 dataset are demonstrated in Fig.~\ref{fig:strands400}. 

\section{Results}

\begin{figure*}[p]
 \centering
   \includegraphics[width=\textwidth]{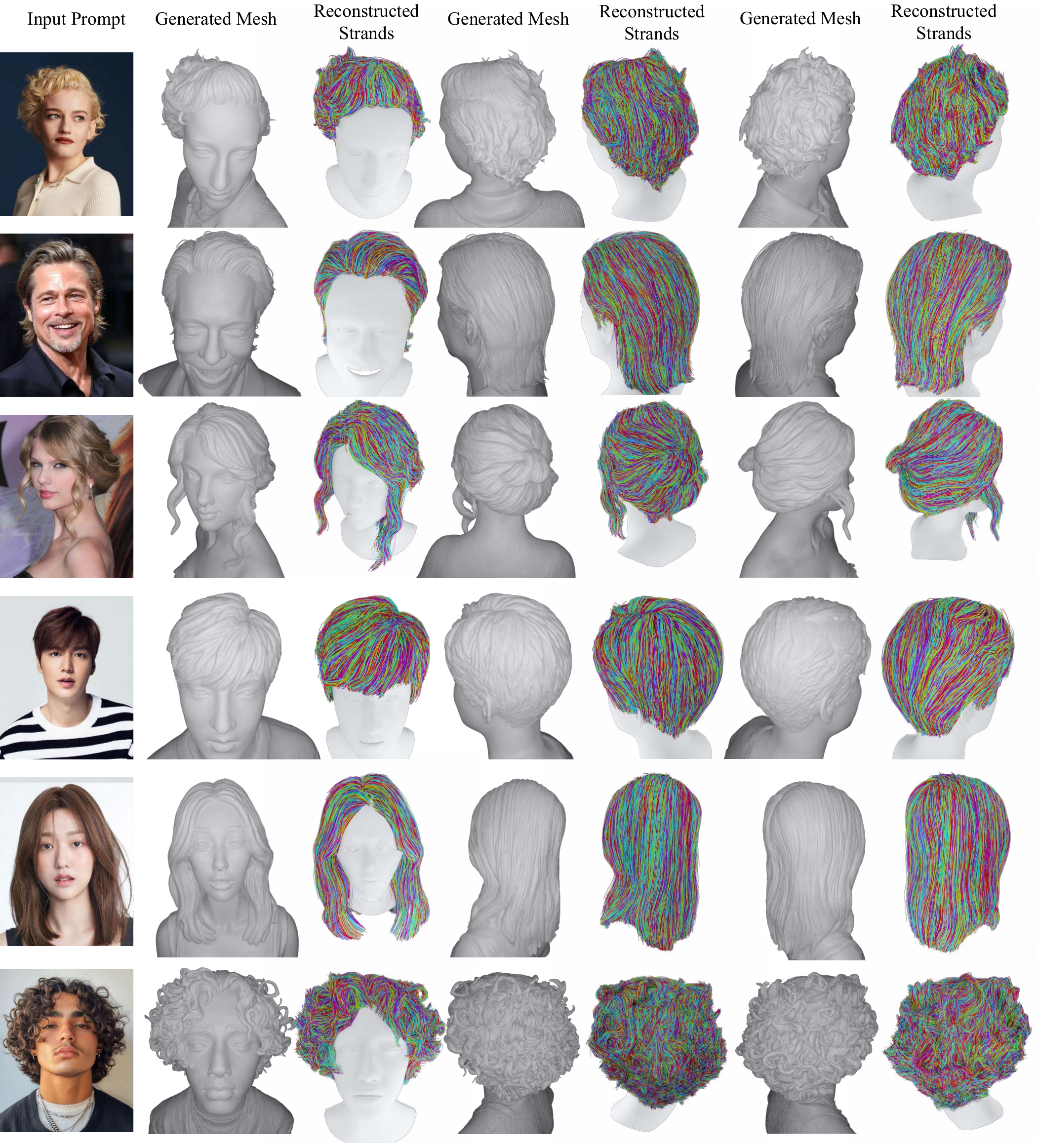}
    \caption{Reconstruction results with meshes from an off-the-shelf image-to-mesh model, demonstrating our method's ability to generate diverse hairstyles from single images.}
    \label{fig:meshy_supp_image}
\end{figure*}
\begin{figure*}[h]
 \centering
   \includegraphics[width=\textwidth]{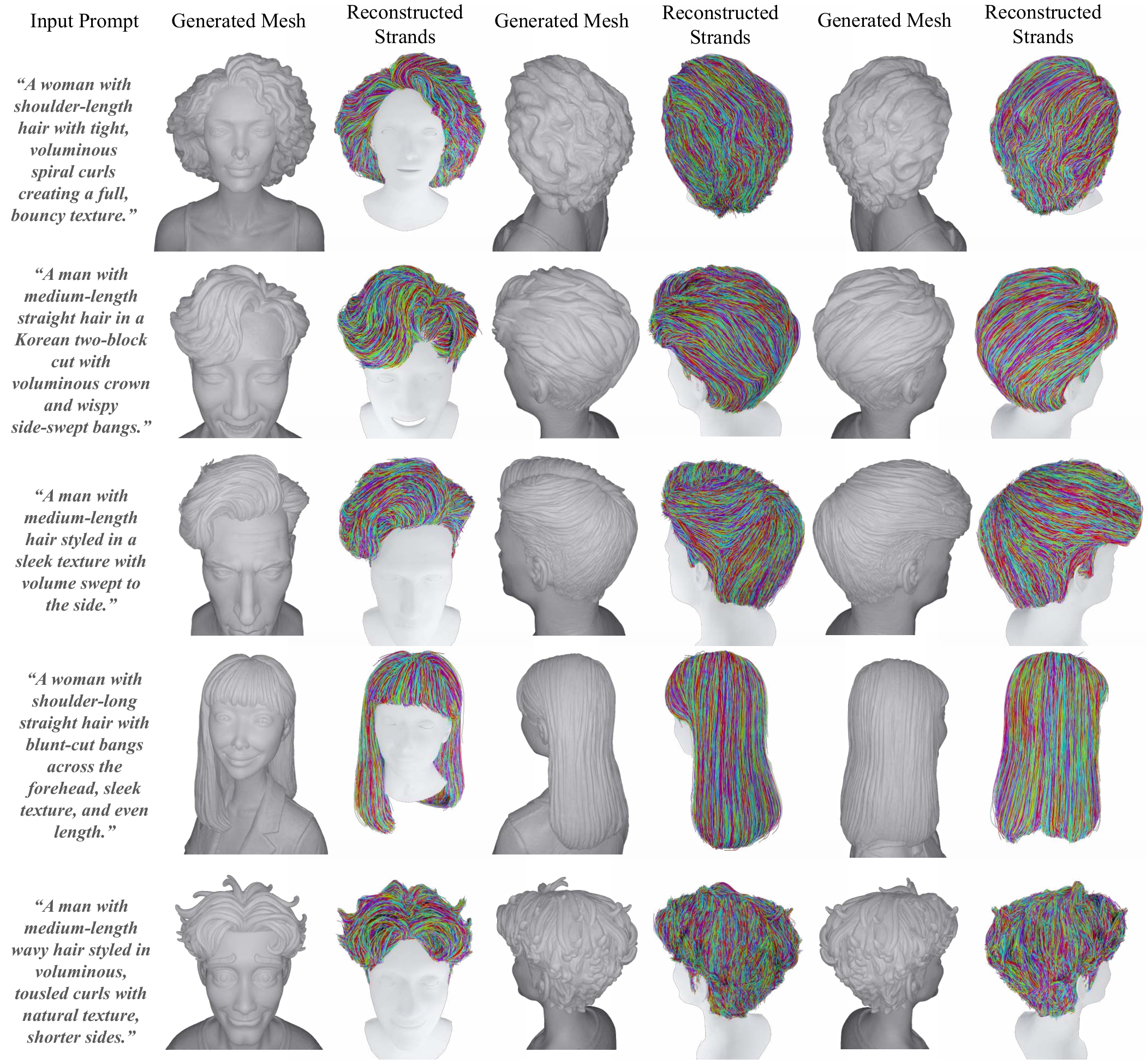}
   \vspace{-0.5cm}
 \caption{Reconstruction results with meshes from an off-the-shelf text-to-mesh model, demonstrating our method's ability to generate diverse hairstyles from text descriptions.}
 \label{fig:meshy_supp_text}
\end{figure*}

\noindent \textbf{Extended application.}
We provide additional results demonstrating our method's versatility with head meshes from off-the-shelf generative models. Figure~\ref{fig:meshy_supp_image} presents extended reconstruction results using an image-to-mesh model, while Figure~\ref{fig:meshy_supp_text} shows results from a text-to-mesh model.

~


~

\begin{figure*}[p]
 \centering
   \includegraphics[width=\textwidth]{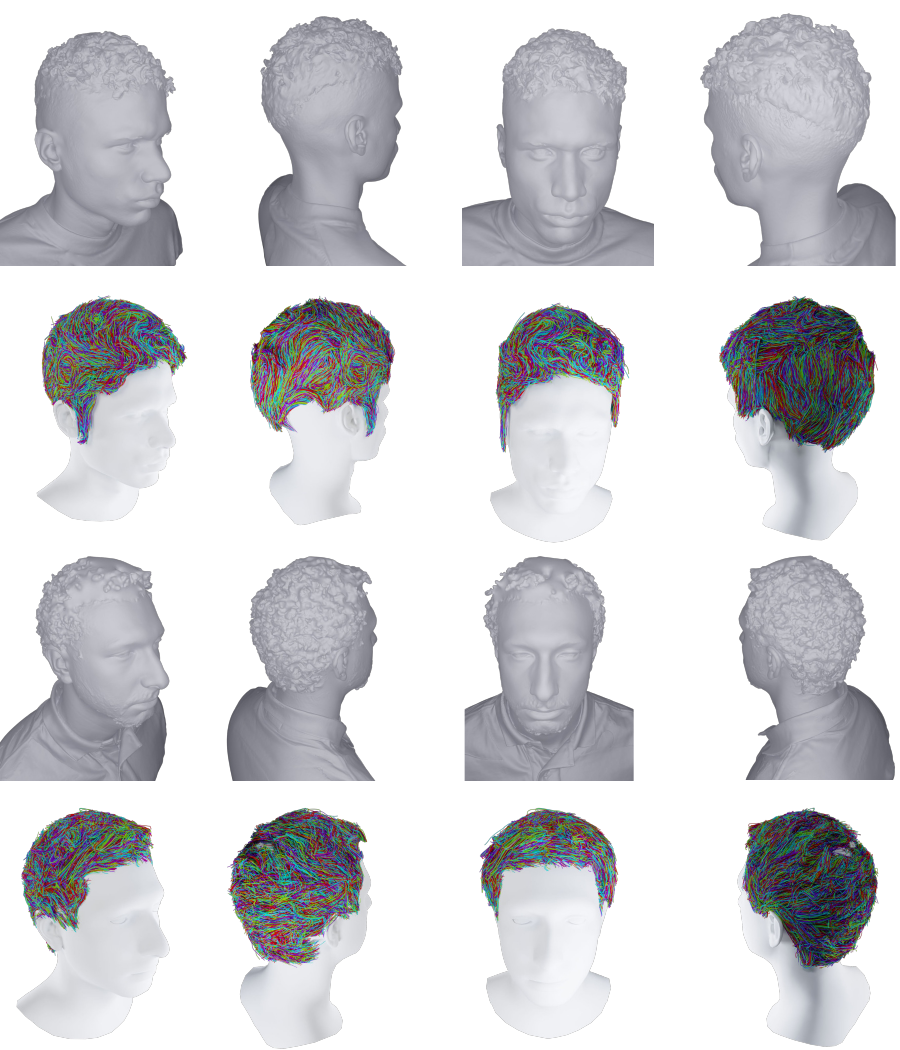}
    \caption{Limitations of our method: reconstruction of very short and curly hairstyles. 
    For very short hairstyles (two top rows), the hair mesh typically lacks sufficient curvature details.
    For curly hairstyles (two bottom rows), the crest lines algorithm fails to capture high-frequency details. }
    \label{fig:failure_cases}
\end{figure*}

\noindent \textbf{Limitations.}
While our method performs well on wavy and straight hairstyles, very curly scenes remain challenging because scans often fail to capture such high-frequency details, amplifying noise in the orientation estimators. 
As an example, we provide two samples where our method fails to reconstruct. Specifically, our method is less robust to curly and very short hairstyles, as illustrated in Fig. \ref{fig:failure_cases}. One approach to address such limitation is by improving the hairstyle diffusion priors.

\end{document}